\documentclass[runningheads]{llncs}
\usepackage[T1]{fontenc}
\usepackage{graphicx}
\usepackage{amsmath}
\usepackage{mathtools}
\usepackage{xspace}
\usepackage{xcolor}
\usepackage{booktabs}
\usepackage{float}
\usepackage{graphicx}
\usepackage{subcaption}
\usepackage[colorlinks=true,
            linkcolor=blue,
            urlcolor=blue,
            citecolor=blue]{hyperref}

\newcommand{\vm}[1]{\boldsymbol{#1}}

\begin{document}
\raggedbottom
\title{Natural Sit-to-Stand Motion Synthesis For Humanoids via Guided Assistance Curricula and Staged Rewards}
\titlerunning{Natural Sit-to-Stand Motion Synthesis}

\author{%
Meet Pal Singh\inst{1} \and
Vyankatesh Ashtekar\inst{1}\thanks{Corresponding author.} \and
Ashish Dutta\inst{1}
}

\authorrunning{M. Pal et al.}

\institute{%
Department of Mechanical Engineering, Indian Institute of Technology Kanpur,
Kanpur, Uttar Pradesh, India\\
\email{\{meetpal22, vyankatesh20, adutta\}@iitk.ac.in}
}
\maketitle
%


\begin{abstract}
A humanoid has infinitely many ways to stand up from sitting while maintaining balance, making sit-to-stand (STS) a challenging control problem. We synthesise natural humanoid STS motion from scratch using reinforcement learning, without demonstrations or reference trajectories. A single Proximal Policy Optimisation policy learns smooth, human-like rising driven by three complementary components. (i)~A coupled force/chair-height curriculum is used. A vertical pelvis-assist force aids early trajectory exploration and decays over training. Taller chairs are unlocked with decaying assisting force. This ensures that the policy masters a viable STS trajectory at each chair height before being exposed to harder ones, avoiding the premature distribution shift that otherwise collapses generalisation. (ii)~Motion robustness is achieved by randomly sampling from a large number of inverse kinematics-generated initial and target poses spanning over eight chair heights. (iii)~A set of rewards is defined inspired from biomechanics and optimal control studies. They shape the robot’s angular momentum for seat-off, and enable support-region transition via centre of pressure attraction function to ensure smooth low-effort actuation. On a deterministic force-free evaluator, the policy attains more than 97\% balanced-standing success across eight chair heights. The policy generalises smooth motion across chair heights and enables the robot to rise from substantially deep-seated postures as compared to the state of the art.
\end{abstract}
\section{Introduction}
Humanoid robots are increasingly being integrated into human environments, making contact-rich whole-body transitions like the sit-to-stand (STS) essential.
This paper synthesises the STS motion for a humanoid robot using reinforcement learning~(RL) from scratch, without imitation or reference-motion tracking.

Collectively, the following four categories encompass the existing
approaches to STS in the literature: (a)~pure
optimisation~\cite{Ozsoy2014DETC,Yang2020,Mungai2021,Aller2022,Howell2022,Zambella2023}, (b)~optimisation with a reference trajectory~\cite{Mistry2010,Fierro2014}, (c)~pure ~RL~\cite{Jiang2025,HuangT-RSS-25}, and (d)~RL with a reference
motion~\cite{liao2025beyondmimic,allshire2025}. The formulation of the STS problem for exoskeletons, see e.g.~\cite{Mungai2021}, closely resembles that of humanoid robots.

Optimisation-based synthesis spans forward-dynamics shooting~\cite{Howell2022,Aller2022}, inverse-dynamics ~\cite{Ozsoy2014DETC,Yang2020}, and direct  collocation~\cite{Mungai2021} methods. The~STS problem has to be  usually split into two~\cite{Aller2022} or more~\cite{Ozsoy2014DETC} domains due to contact discontinuities. 
Algorithms from MuJoCo~MPC~\cite{Howell2022} demonstrated real-time motion synthesis but yielded spasmodic humanoid~STS. It remains a prominent yet under-explored alternative to~RL.
A reference trajectory shrinks this search space, but kinematic and dynamic
disparity means the recorded motion must be retargeted to the robot.
Mistry~et~al.~\cite{Mistry2010} guided a planar~STS using human centre of
mass~(COM) and centre of pressure~(COP) trajectories, mapping motion to a scaled
skeleton and solving joints by inverse kinematics~(IK).
Gonz\'alez-Fierro~et~al.~\cite{Fierro2014} fitted a Gaussian
zero-moment-point~(ZMP)/effort reward by differential evolution and later surpassed the human baseline through skill innovation.

Optimisation based methods require smooth gradient of objective function, unavailable in contact-rich motions. The motion resulting from~RL policy, say Proximal Policy Optimisation~(PPO), is based on a large number of sampled returns, avoiding the need for analytical derivatives of contact dynamics.
Huang et al.~\cite{HuangT-RSS-25} demonstrated a learning-based approach for humanoid stand-up in a wide range of scenarios. They used a multi-critic architecture with separate task, style, regularisation, and post-task reward groups, a vertical pull-force exploration curriculum, an action-bound rescaler that limits motion speed, and smoothness regularisation. However, the resulting STS motion still involved abrupt movements.
Jiang et al.~\cite{Jiang2025} formulated STS and stand-to-sit using~RL from scratch, and demonstrated a shallow STS motion on a physical robot. They had to ensure a favourable sitting zone to ensure good standing up success, otherwise the success rate dropped by~40-60\%~\cite{Jiang2025}.
Imitation learning mimics contact-rich human motions, inheriting their physical
consistency. Allshire~et~al.~\cite{allshire2025} showed shallow STS among many
skills, but the motions were abrupt, lacking smooth human-like transitions.

State-of-the art methods using~RL to solve STS have two clear shortcomings. First, the
demonstrated initial seated postures (see e.g.,~\cite{Jiang2025,allshire2025}) are
shallow in the sense that the pelvis rests on the seat edge with inclined thigh links,
such that the majority of the robot's weight is already on its feet. Second,
learned movements are typically abrupt if the robot sits deep in a
chair~(e.g.~\cite{HuangT-RSS-25}). A smooth, controlled, energy-efficient transition like humans is desired to
ensure the safety of the robot and its surroundings.


The main contributions of this work are as follows. We design a pure~RL-based
STS motion synthesis method that uses physics-based rewards inspired from the
optimal control literature to obtain a human-like stand-up. For sample-efficient and physically consistent learning, we systematically
randomise the initial seated and target standing configurations with whole-body~IK
over a large pose library spanning eight chair heights. A dimensionless rise-fraction unifies staged rewards across varying chair geometries, while a coupled curriculum linking pelvis assistance force decay to progressive chair-height unlocking drives robust motion discovery and refinement. The result is a smooth, balanced motion
amenable to physical implementation.
\section{Methodology}
In traditional optimisation frameworks, the STS problem is set up as follows. The robot's initial and final configurations are usually predefined. The cost function aggregates quantities integrated over the motion, such as the magnitude of actuator efforts, the rate of change of effort, mechanical work done by the
actuators, equal distribution of ground reaction forces between the two feet~\cite{Ozsoy2014DETC}, and attractors pulling the ZMP~\cite{vukobratovic2004} or the COP toward a desired support location. The constraints include joint angle, velocity, acceleration and torque limits, contact constraints at the
foot--ground and seat interfaces, a requirement that the net COP lie within a preferred zone of the support region, and STS stage durations~\cite{Ozsoy2014DETC}. A robot is balanced only if its ZMP/COP lies strictly inside the support polygon at all times.
Depending on whether forward-dynamics shooting, inverse dynamics, or direct collocation is used, the actuator efforts, joint trajectories, or both are parametrised as smooth splines.
%
\subsection{STS Formulation using RL}
We cast STS as a finite-horizon Markov decision process (MDP)
$\langle S,A,T,R,\gamma\rangle$ and solve it with~PPO~\cite{schulman2017}. The chosen robot is the 29 degree-of-freedom (DOF)
Unitree~G1 modelled in MuJoCo~\cite{todorov2012}. The policy acts at $50$~Hz (frame-skip $10$ over a $2$~ms integrator step) and each episode runs up to $40$~s ($2000$ steps), ending early on a safety termination (the COM dropping below a floor, or trunk roll/pitch
beyond limits).

\paragraph{Observation and action spaces.}
The policy is proprioceptive. At step $t$ it observes the $97$-dimensional vector
\begin{align}
\vm{o}_t = [\,\vm{g}_b,\ \vm{\omega}_b,\ \vm{a}_b,\ \vm{q},\ \dot{\vm{q}},\
\vm{a}_{t-1},\ \beta_{\max}\,],
\end{align}
where $\vm{g}_b\in\mathcal{R}^3$ is the gravity vector projected into the torso
frame, $\vm{\omega}_b\in\mathcal{R}^3$ and $\vm{a}_b\in\mathcal{R}^3$ are the torso
angular velocity and linear acceleration from the inertial measurement unit (IMU),
$\vm{q},\dot{\vm{q}}\in\mathcal{R}^{29}$ are the joint positions and velocities,
$\vm{a}_{t-1}$ is the previous action, and $\beta_{\max}$ is the scalar
action-bound multiplier (Sec.~\ref{sec:curr}). The action
$\vm{a}_t\in[-1,1]^{29}$ is an incremental (delta) joint command, mapped to a
proportional-derivative (PD) target and the applied joint torque by the global action rescaler $\beta_{\max}$,
\begin{align}
\vm{u}_t &= \mathrm{clip}\big(\vm{q}_t + \vm{a}_t\odot\beta_{\max},\,
\vm{q}_{\min},\,\vm{q}_{\max}\big),\\
\vm{\tau}_t &= \vm{K}_p\,(\vm{u}_t-\vm{q}_t)-\vm{K}_d\,\dot{\vm{q}}_t,
\end{align}
where $\odot$ is the element-wise product and $\vm{K}_p,\vm{K}_d$ are the servo
gains. The rescaler $\beta_{\max}\le1$ shrinks the admissible per-step excursion.
Training uses vectorised environments with observation and reward normalisation.

\subsection{Randomisation, Rewards, and Curriculum}
In~RL literature, adding joint-space noise for initial pose randomisation is
common, but in contact-rich tasks it either produces invalid poses, e.g., bodies
clipping through the environment, or requires severely restricting the noise. To
ensure physically consistent state randomisation, we employ~IK, an approach
scarcely utilised in RL literature. The following DOFs are randomised: the seat
height/final stance height, the torso's position along the seat plane and its
pitch orientation, as well as the planar position of the feet, as depicted in~Fig.~\ref{fig:combined}\subref{fig:IK_randomisation}. The arm DOFs are
randomised in joint space. 
To maintain run-time efficiency, a large number of
poses are generated beforehand and chosen at random at the start of episodes.
The resulting library of poses holds
$29{,}260$ seated states over eight chair heights, paired with a
standing-reference pool, so that the policy never memorises a single trajectory. From
the seated pose to full standing the head must rise by $0.24$--$0.34$~m which is largest
for the lowest (default) chair and decreasing as the seat is raised.
%
%
\begin{table}[t]
\centering
\caption{Assist-force and chair-height curriculum. Each \emph{lane} is a discrete seat height (lane~0 lowest). New lanes are progressively unlocked as $F_{\max}$ decays.}
\label{tab:curriculum}
\footnotesize
\begin{tabular}{@{}ll@{}}
\toprule
Assist force range & Unlocked chair lanes\\
\midrule
$F_{\max}>100$~N & \{0\}\\
$80<F_{\max}\leq100$~N & \{0,1\}\\
$60<F_{\max}\leq80$~N & \{0,1,2,3\}\\
$40<F_{\max}\leq60$~N & \{0,1,2,3,4,5\}\\
$20<F_{\max}\leq40$~N & \{0,1,2,3,4,5,6,7\}\\
$F_{\max}\leq20$~N & \{1,2,3,4,5,6,7\}\\
\bottomrule
\end{tabular}
\end{table}
%
%
%
\paragraph{Rewards.}
A common term applies every step; exactly one of three rise-stage terms is then
added based on $\rho$ (Table~\ref{tab:rewards}).

\emph{Rise-fraction.} Different chairs place the seated head at different heights,
so absolute thresholds would misclassify postures. At reset we record the seated
head height
$h_0$ and use
$\rho=(h_{\mathrm{head}}-h_0)/\max(h_{\mathrm{stand}}-h_0,\epsilon)\in[0,1]$
($h_{\mathrm{stand}}=1.3236$~m), giving stage boundaries (rise-low $\rho<0.1$,
rise-mid $\rho<0.8$, standing $\rho\ge0.8$) that are identical across chairs.

\emph{COP shaping.} The COP is the force-weighted mean of nine contact sensors
(four per foot, one at the pelvis/seat), with a foot-midpoint fallback in flight.
Rising terms pull the COP from seat toward feet; standing adds a ZMP-style pull to
the foot-support-hull centroid for a centred, tip-free stance. This relies only on
measured contact forces rather than privileged balance signals~\cite{Poddar2025}.

\emph{Angular momentum shaping.} A signed bonus $\omega_y\lvert\omega_y\rvert$ in rise-low
rewards forward pitch rate for seat-off; a standing term $\exp(-5\,\omega_y^2)$
damps it; a roll/yaw-rate penalty acts throughout for lateral stability.

\emph{Anti-parking.} The penalty $-n_b$ counts contacts between the environment
(ground or seat) and any non-foot body (calf, thigh, pelvis). Further Gaussian terms
reward uprightness, flat feet, symmetry, and a per-episode standing reference,
with smoothness penalties on action rate, acceleration, joint speed, and
power~\cite{HuangT-RSS-25,Poddar2025}. These terms are inspired by the physics of
human STS in classical biomechanics---forward angular momentum to break seat contact,
support-region regulation through the COP/ZMP, and smooth low-effort
actuation.
\begin{table}[t]
\centering
\caption{Rewards and their weights. Notation: $h_s{=}1.3236$~m, $c$:\,COP,
$\bar c_{\mathrm{hull}}$: support-hull centroid, $\tau_{\mathrm{f}}$: foot tilt,
$n_b$: non-foot brace contacts, $e_{\mathrm{sym}}$: mirror error,
$\Delta a{=}a_t{-}a_{t-1}$, $\Delta^2 a{=}a_t{-}2a_{t-1}{+}a_{t-2}$, other symbols
as in the text). $^\dagger$~rise-low ($\rho<0.1$) only, with
$c_{\mathrm{ref}}{=}\tfrac12(c_b+\bar c_{\mathrm{feet}})$ there and the hull
centroid otherwise.}
\label{tab:rewards}
\footnotesize
\begin{tabular*}{\linewidth}{@{\extracolsep{\fill}}llr@{}}
\toprule
\textbf{Term} & \textbf{Expression} & \textbf{Weight}\\
\midrule
\multicolumn{3}{@{}l}{\textit{Common} (applied every step)}\\
roll              & $\exp(-10\phi^2)$ & $1$\\
waist yaw         & $\exp(-10q_{\mathrm{w}}^2)$ & $1$\\
foot contact      & $\exp(-50(d_\ell+d_r))$ & $1$\\
spin              & $\exp(-0.5(\omega_x^2+\omega_z^2))$ & $1$\\
symmetry          & $\exp(-5\lVert e_{\mathrm{sym}}\rVert^2)$ & $1$\\
ankle flat        & $\exp(-10\tau_{\mathrm{f}})$ & $1$\\
survival          & $1$ & $1$\\
smoothness        & $-0.1\lVert\Delta a\rVert-0.1\lVert\Delta^2 a\rVert-10^{-3}\lVert\dot q\rVert-2.5{\times}10^{-4}\lVert\tau\odot\dot q\rVert$ & $1$\\
anti-brace        & $n_b$ & $-1$\\
\midrule
\multicolumn{3}{@{}l}{\textit{Rising} ($\rho<0.8$)}\\
progress          & $\mathrm{clip}(300\,\Delta h,-1,5)$ & $2$\\
head height       & $\exp(-20(h-h_s)^2)$ & $6$\\
COP centring      & $\exp(-15\lVert c-c_{\mathrm{ref}}\rVert^2)$ & $2$\\
pitch-rate$^\dagger$ & $\omega_y\lvert\omega_y\rvert$ & $0.5$\\
\midrule
\multicolumn{3}{@{}l}{\textit{Standing} ($\rho\ge0.8$)}\\
head height       & $\exp(-30(h-h_s)^2)$ & $6$\\
torso pitch       & $\exp(-20\theta^2)$ & $4$\\
ZMP               & $\exp(-15\lVert c-\bar c_{\mathrm{hull}}\rVert^2)$ & $4$\\
pose              & $\exp(-10\lVert q-q^{\mathrm{ref}}\rVert^2)$ & $2$\\
pitch damp        & $\exp(-5\omega_y^2)$ & $1$\\
COM velocity      & $\exp(-5\lVert v_{xy}\rVert^2)$ & $1$\\
\bottomrule
\end{tabular*}
\end{table}
%
\paragraph{Curriculum and training.}\label{sec:curr}
Two curricula run together. \emph{Assist decay:} the pelvis helper force
$F_{\max}$ and the action bound $\beta_{\max}$ start at $200$~N and $1.0$ and decay
on each pass of a behavioural gate. The gate uses two levels of averaging: per
episode we take the mean head height over the final $50$ steps of an episode (a sustained,
end-of-episode height, not a peak), and the gate passes once at least $60\%$ of the
last $35$ completed episodes exceed the fixed head-height gate which is $85\%$ of the base-chair rise. Each pass lowers $F_{\max}$
by $20$~N  and $\beta_{\max}$ by
$0.02$. \emph{Chair unlocking:} as $F_{\max}$ decays, taller
chairs are added per Table~\ref{tab:curriculum}, so a new height band appears only
after the robot reliably stands at the current difficulty thus avoiding the abrupt
distribution shift that otherwise collapses generalisation.
\section{Results and Discussion}
We evaluate the trained policy in simulation across all eight chair heights, reporting success, smoothness, safety, balance, and single-component ablations. The results are compiled in a video\footnote{\url{https://youtu.be/SgSlgJRrlcE}} and code to be open sourced on GitHub\footnote{\url{https://github.com/meetpal644/}}.

All policies are scored with one deterministic, force-free evaluator over the eight chair lanes ($50$ resets per lane), following~\cite{Jiang2025,HuangT-RSS-25}. \emph{Action jitter} and \emph{DoF jitter} are the mean per-step changes~$\tfrac1T\sum_t\lVert a_t-a_{t-1}\rVert$ (dimensionless) and $\tfrac1T\sum_t\lVert~q_t-q_{t-1}\rVert$~(rad); the safety coefficients~$S_\tau$ and~$S_q$ are the fractions of steps with all joint torques and all joint angles respectively within their limits. A trial is a \emph{balanced success} only if, over the final second, the head stays above the gate, horizontal COM
speed $<0.15$~m/s, and trunk tilt $<0.4$~rad; a fall before  timeout is a failure. Success/fall use all $50\times8=400$ trials; other means exclude early-terminated trials.

\begin{figure}[h]
    \centering
    \begin{subfigure}[t]{0.6\textwidth}
        \centering
        \includegraphics[height=3.5cm]{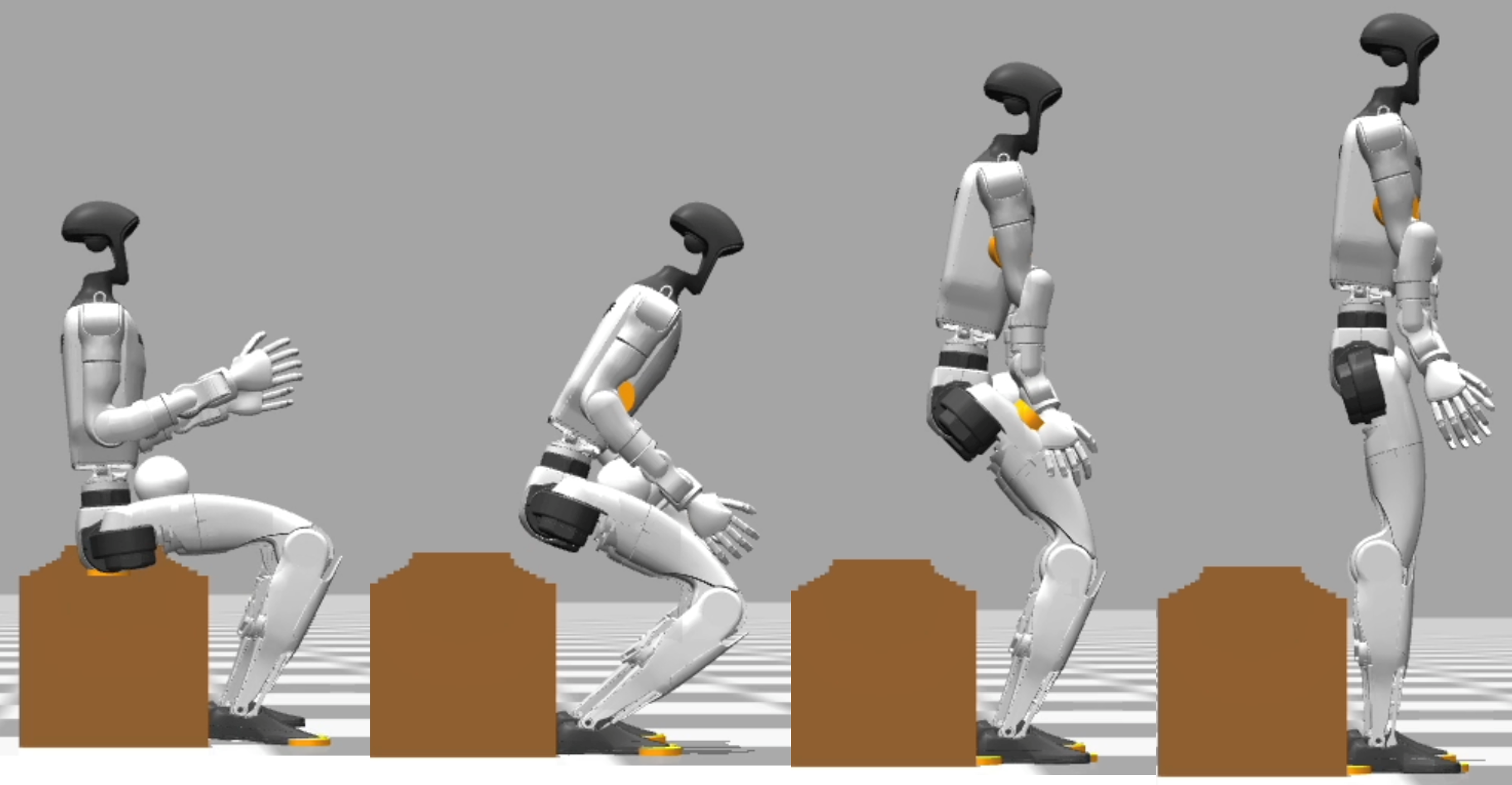}
        \caption{Stop-motion frames captured (sagittal plane) of the smooth and natural learned STS motion . Refer to the \href{https://youtu.be/SgSlgJRrlcE}{video} for full motion visualisation.}
        \label{fig: sts_stopmotion}
    \end{subfigure}
    \hfill
    \begin{subfigure}[t]{0.3\textwidth}
        \centering
        \includegraphics[height=3.5cm]{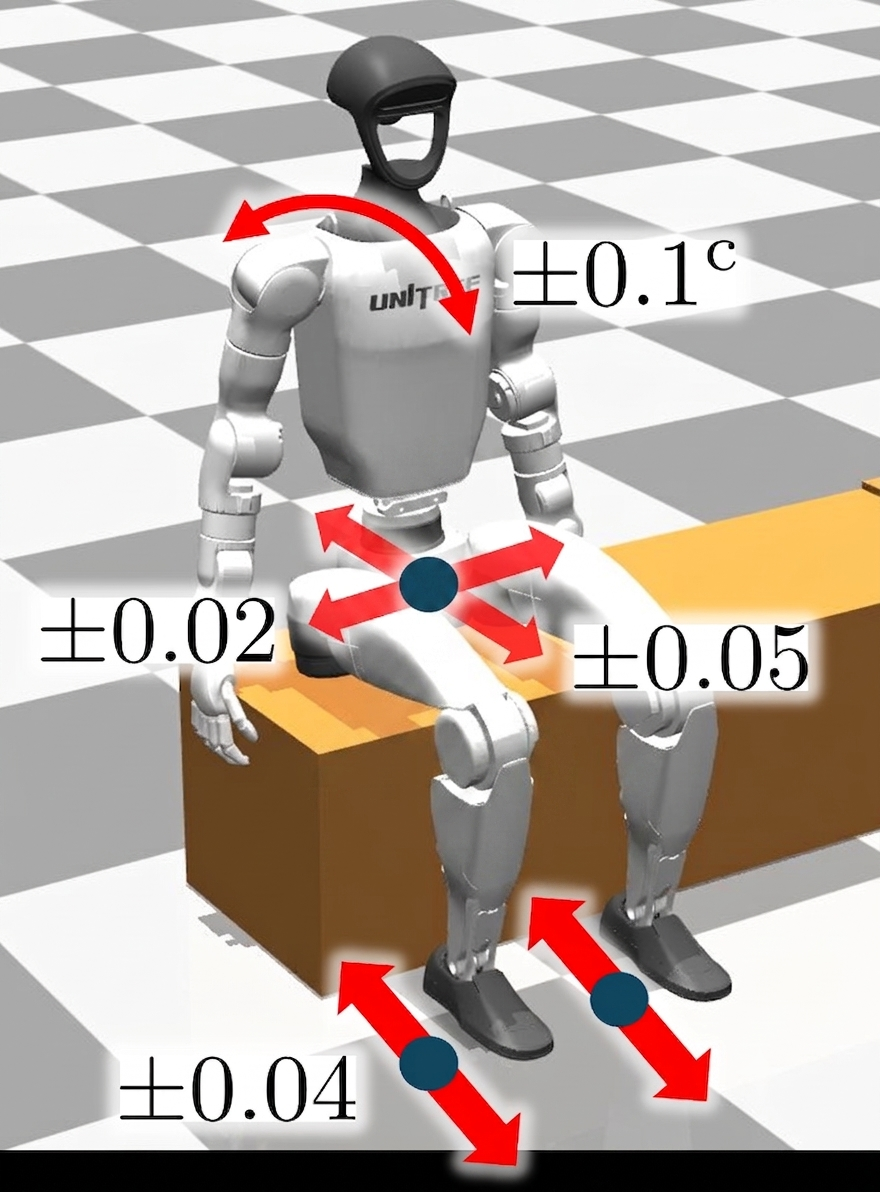}
        \caption{Random initial poses generated for each seat height using IK.}
        \label{fig:IK_randomisation}
    \end{subfigure}
    \caption{Learned STS motion and randomized initial poses.}
    \label{fig:combined}
\end{figure}

\begin{figure}[h]
\centering
    \begin{subfigure}[t]{0.45\textwidth}
    \centering
        \includegraphics[height=3cm]{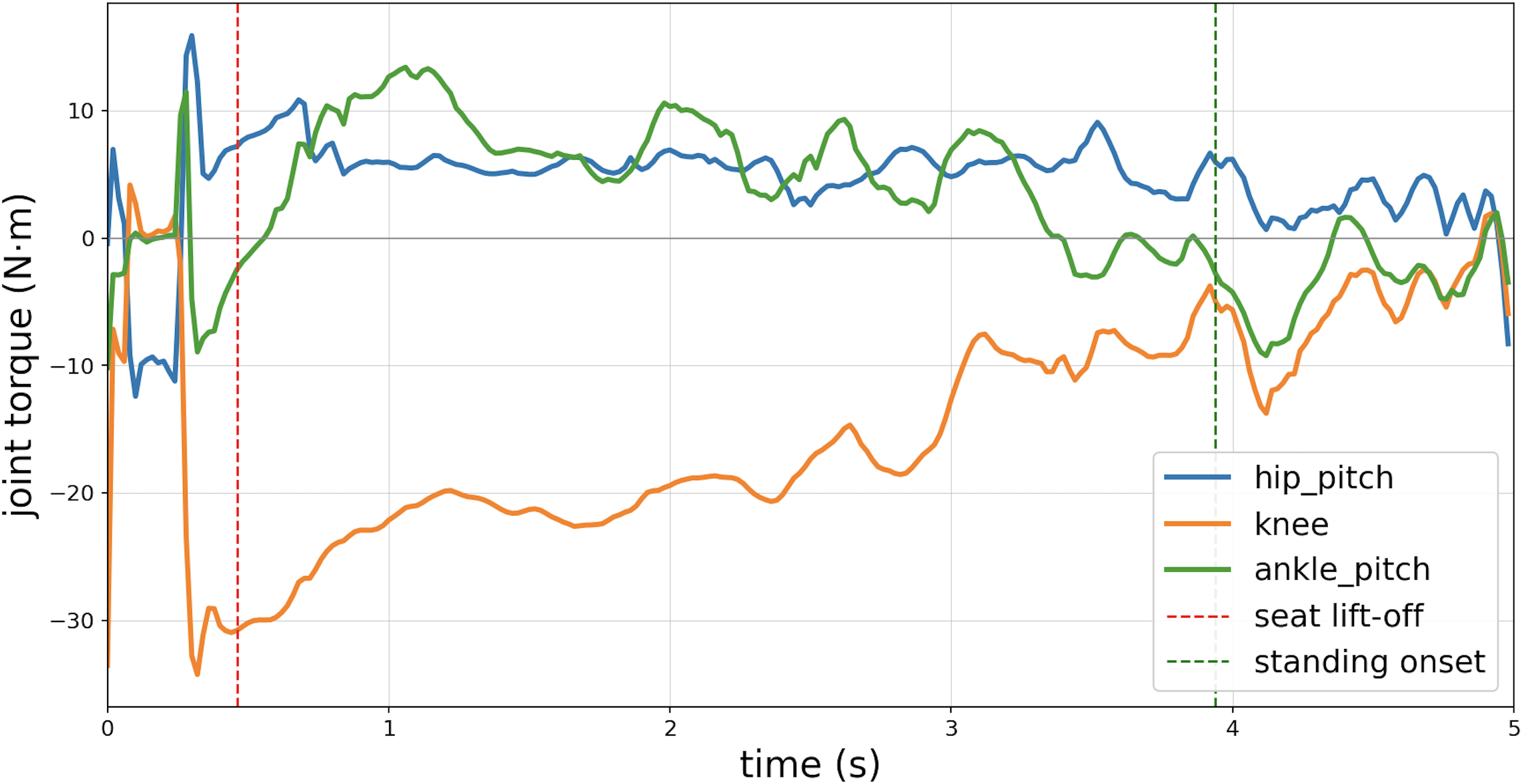}
        \caption{}
        \label{fig:results_jt_trq}
    \end{subfigure}
    \hfill
    \begin{subfigure}[t]{0.45\textwidth}
    \centering
        \includegraphics[height=3cm]{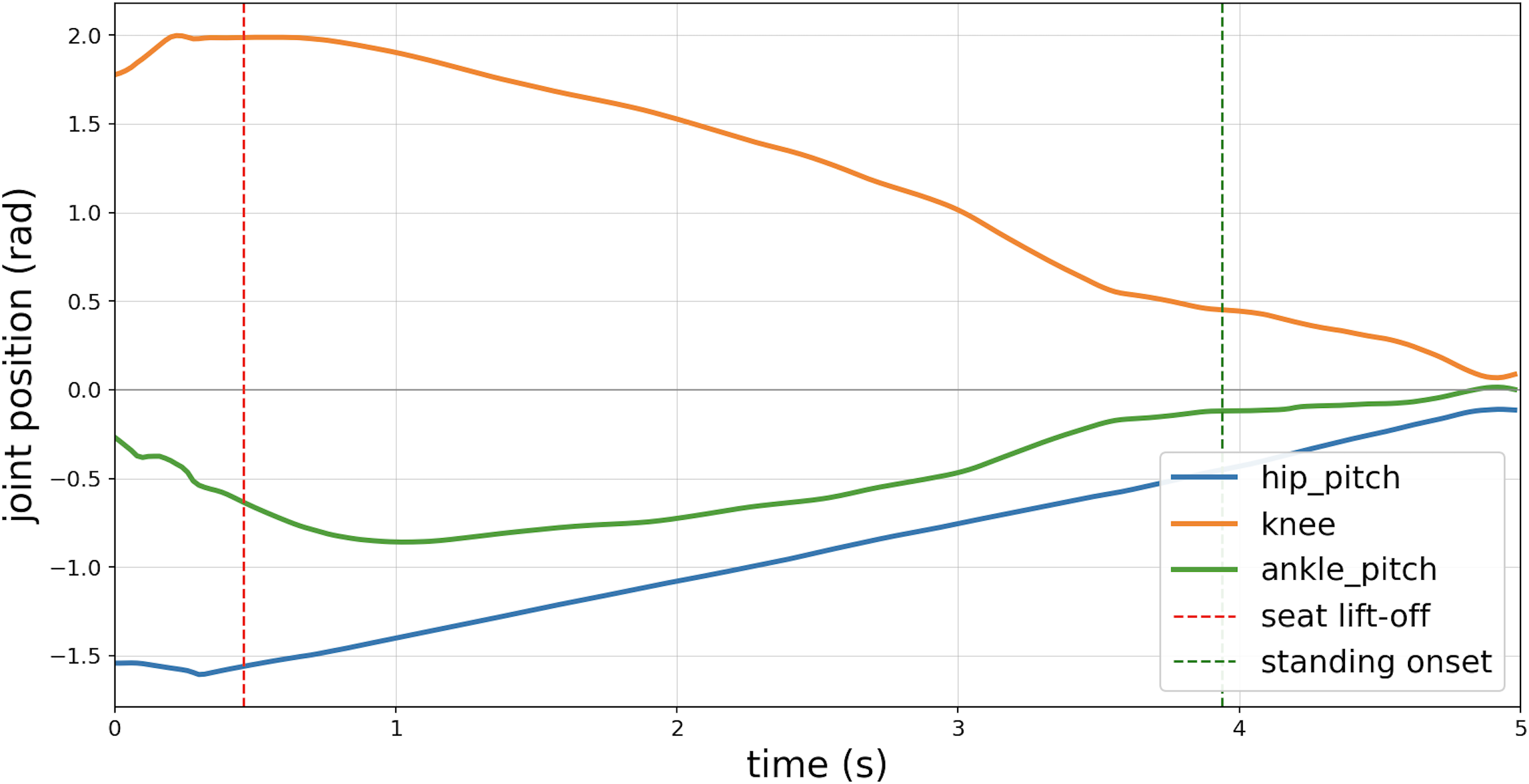}
        \caption{}
        \label{fig:results_jt_ang}
    \end{subfigure}
\caption{Temporal variation of~(\subref{fig:results_jt_trq}) pitch joint torques and (\subref{fig:results_jt_ang})~angles during the rising phase. The red vertical line~($\approx0.5$~s) indicates seat-off while the green dashed vertical line~($\approx3.9$ s) marks 90\% rise completion}
\label{fig:angle_trq_plot}
\end{figure}

\begin{figure}[h]
\centering
\includegraphics[width = 0.65\textwidth]{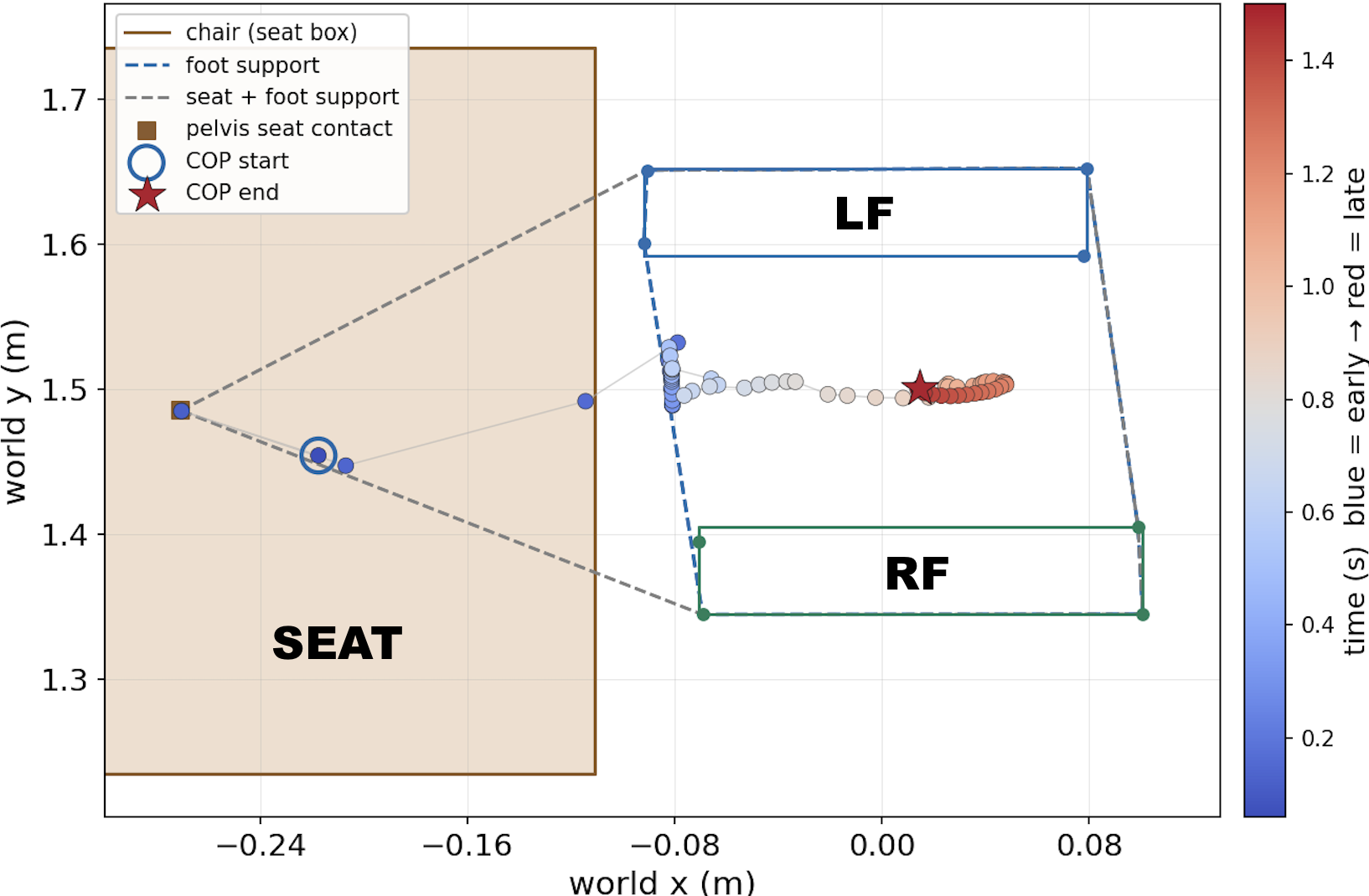}
\caption{The robot transfers its balance
(quantified via its~COP location) from the chair (encircled point) to the centroid of the foot-only support region.}
\label{fig:cop_locus}
\end{figure}

On the deterministic, force-free evaluation the full policy attains
\textbf{97.8\%} balanced-standing success across the eight chair lanes, with mean
ascent time $t_{\mathrm{stand}}=3.54$~s and energy consumption $115.6$~J
(Table~\ref{tab:ablation}). Figure~\ref{fig:combined}\subref{fig: sts_stopmotion} depicts a few frames from a representative
rollout which shows the robot rising up naturally from a deep-seated posture. Figure~\ref{fig:angle_trq_plot}\subref{fig:results_jt_trq} indicates joint torque activity~(within actuator bounds and consistent with the perfect torque-safety score ($S_\tau=1.00$)) to produce enough angular momentum rate that results in smooth sagittal-plane joint
trajectories~ (Fig.~\ref{fig:angle_trq_plot}\subref{fig:results_jt_ang}). 
It can be observed in the video that the arm movements, though not explicitly commanded, synergistically contribute to the angular momentum rate required for the transition.
Figure~\ref{fig:cop_locus}
adds the contact view which depicts the COP begins near the pelvis--seat contact~(indicating a deep seating posture), moves into
the foot-only support polygon, and settles at the centroid of the same, as the weight transfers from seat to feet.
\begin{table}[!t]
\centering
\caption{Ablation study: each ablation omits one system component. All policies are evaluated deterministically (force-free, 8 lanes, 50 resets per lane).}
\label{tab:ablation}
\footnotesize
\setlength{\tabcolsep}{3pt}
\resizebox{\textwidth}{!}{%
\begin{tabular}{@{}llcccccccc@{}}
\toprule
Run & Removed component & Success\,(\%) & Fall\,(\%) & Act.\,jit. & DoF\,jit. & $S_\tau$ & $S_q$ & Energy\,(Joules) & $t_{\mathrm{stand}}$\,(s)\\
\midrule
Full          & none (reference)          & \textbf{97.8} & 2.2 & 1.535 & 0.105 & 1.00 & 0.963 & 115.6 & 3.54\\
Rise-frac-off & rise-fraction staging     & 0.0 & 59.3 & 2.522 & 0.059 & 1.00 & 0.187 & 94.80 & N/A\\
COP-off       & measured-COP shaping      & 97.0 & 3.0 & 3.35 & 0.103 & 1.00 & 0.990 & 115.2 & 3.60\\
Schedule-off  & progressive chair unlock  & 85.0 & 15.0 & 1.393 & 0.073 & 1.00 & 0.807 & 116.9 & 3.50\\
Omega-off     & sagittal pitch-momentum   & 88.0 & 11.8 & 0.64 & 0.045 & 1.00 & 0.882 & 115.5 & 3.66\\
\bottomrule
\end{tabular}%
}
\end{table}
\paragraph{\textit{Ablation Study :}}
The ablations show that chair-height generalisation needs both reward
normalisation and gradual widening of the start-state distribution. Without
rise-fraction staging, absolute-height phases do not transfer across seats. In
the schedule-off run, all chair lanes are sampled from the start while the
assist-force/action-bound curriculum still decays; the reduced success indicates
that progressive unlocking helps the policy retain stable base-chair behaviour
while adding taller seats. Removing sagittal pitch-momentum lowers success, while
removing COP shaping preserves headline success but more than doubles action
jitter, indicating that measured-COP terms mainly regularise load transfer and

\paragraph{\textit{Limitations :}}
Two limitations remain. When heels start close to the seat edge (2--3~cm clearance), the policy occasionally braces against the seat to balance, yielding a less natural rise. The STS motion slows down towards the end: the bulk of the rise ($\approx 90\%$) completes in 3--4~s, with full settling taking up to 5~s.
\section{Conclusion}
Results demonstrate that a smooth and human-like deep-seated humanoid STS motion can be learned by pure reward-shaped RL.
An IK-generated multi-seat-height pose library, rise-fraction staging, COP and angular momentum shaping, and coupled force/action-bound chair curriculum produce a single
proprioceptive PPO policy with more than 97\% deterministic success over various chair heights. 
Ablations identify rise-fraction and progressive unlocking of chair heights as key generalisation drivers. Future work will target sim-to-real validation.
%
\section*{Acknowledgements}
The support and the resources provided by PARAM Sanganak under the National
Supercomputing Mission, Government of India at the Indian Institute of Technology,
Kanpur are gratefully acknowledged.

%

%
%
\bibliographystyle{splncs04}
\bibliography{sts_ref.bib}
\end{document}